# The Comparison of Individual Cat Recognition Using Neural Networks

Mingxuan Li[1] Kai Zhou[2*]

[1] Shanghai Pinghe School, No.333 Shenqi Road, Pudong New Area, Shanghai, Shanghai 201208,China

[2] Department of Neurology of the Second Affiliated Hospital, Interdisciplinary Institute of Neuroscience and Technology, Zhejiang University School of Medicine, Zhejiang University, Hangzhou 310000, China

Address correspondence to Kai Zhou, Department of Neurology of the Second Affiliated Hospital, Interdisciplinary Institute of Neuroscience and Technology, Zhejiang University School of Medicine, Zhejiang University, Hangzhou 310000, China.  Email: 0624684@zju.edu.cn.

## Abstract

Facial recognition using deep learning has been widely used in social life for applications such as authentication, smart door locks, and photo grouping, etc. More and more networks have been developed to facilitate computer vision tasks, such as ResNet, DenseNet, EfficientNet, ConvNeXt, and Siamese networks. However, few studies have systematically compared the advantages and disadvantages of such neural networks in identifying individuals from images, especially for pet animals like cats. In the present study, by systematically comparing the efficacy of different neural networks in cat recognition, we found traditional CNNs trained with transfer learning have better performance than models trained with the fine-tuning method or Siamese networks in individual cat recognition. In addition, ConvNeXt and DenseNet yield significant results which could be further optimized for individual cat recognition in pet stores and in the wild. These results provide a method to improve cat management in pet stores and monitoring of cats in the wild.

## 1. Introduction

Over the past few decades, machine learning has been widely applied to facial recognition, which is used for various purposes such as biometric verification for smartphones (Ríos-Sánchez et al., 2020), security systems (Phawinee et al., 2021), photo grouping, and surveillance (Singh et al., 2023). Deep learning is the key technique for facial recognition, allowing computational models composed of multiple processing layers to learn representations of facial information at multiple levels of abstraction (LeCun et al., 2015). In addition, the models learn to detect intricate structures in large facial datasets by using the backpropagation algorithm. This method tells a model how to change its internal parameters, which are used to compute the representation in each layer from the previous layer (LeCun et al., 2015). In some cases, the classification of faces using deep learning has achieved similar efficiency or outperformed the human brain (Phillips & O'Toole, 2014).

To better complete computer vision tasks with higher efficiency, many network models have been developed, such as ResNet (He et al., 2016), DenseNet (Huang et al., 2017), EfficientNet (Tan & Le, 2019), ConvNeXt (Liu et al., 2022), Siamese networks (Chicco, 2021). During object recognition, different models have intrinsic advantages and disadvantages. However, few studies have systematically compared the advantages and disadvantages of such neural networks in identifying individuals from images, especially for pet animals like cats.

Cats are a common house pet, with 26% of American households owning at least one cat, totaling 65.8 million cats (Applebaum, 2023). Cat owners could benefit from applications of cat recognition through appliances such as smart feeding bowls and smart litter boxes. These products could cater services to individuals and record important health data. Stray cats are also common in areas such as the United States and China, with an estimated 60 to 100 million stray cats in the United States alone (Jessup, 2004). As a predatory invasive species, cats pose a threat to the ecosystem, causing the extinction of at least 20 species of native mammals in Australia (Doherty et al., 2017). The control of cats both in the wild and in pet stores and shelters brings a great burden to human sociality, and machine recognition of individuals may be the first step for improving cat management. In the present study, we systematically compared the efficacy of different convolutional neural networks in cat recognition. We found traditional Convolutional neural networks (CNNs) trained with transfer learning have better performance than models trained with a fine-tuning method or Siamese networks in this task. This study may provide insight into how home appliances, management in shelters, and cat monitoring in the wild can be improved for the benefit of society.

## 2. Related Work

Previous, there has been limited research on using deep learning methods to achieve facial identification in animals such as cats. Adam Klein used EfficientNet with regularized triplet loss to generate 64-dimensional embeddings for cat face verification and identification. They report 95% accuracy on the verification task and 81% on the rank-5 identification task (Klein, 2019). Alexy Dosovitskiy uses a ResNet50 Siamese network with KNN and statistical learning methods to achieve 94.4% accuracy for open-set animal facial recognition, including cats and dogs (Alexey Dosovitskiy, 2020 ). However, none of these papers systematically compare the performance of different neural network architectures in accomplishing individual cat recognition. They also limit the recognition of individuals to facial recognition, omitting bodies.

Convolutional neural networks are a dominant paradigm in the field of computer vision and are commonly used for computer vision tasks, especially classification, and achieve state of the art in human facial recognition (Liu et al., 2023; Shah, 2023). Therefore, CNNs may provide a good architectural candidate for the backbone of the identification of cat individuals. Vision transformers (ViT) such as the Swin Transformer have reached state-of-the-art performance in multiple benchmarks, outperforming CNNs (Alexey Dosovitskiy, 2020 ; Liu et al., 2021). However, ViTs require large training datasets to obtain their high performance, hence being unsuitable for the task at hand. Here, we review several networks including ResNet, DenseNet, EfficientNet, ConvNeXt, and Siamese network.

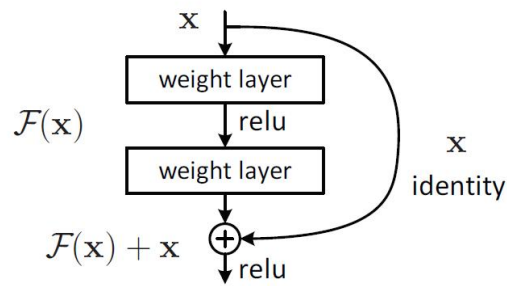

**Figure 1. ResNet model (https://doi.org/10.1109/Cvpr.2016.90)**

**ResNet** (Residual Neural Network). ResNet is a standard CNN architecture that was designed to address the vanishing gradient problem. ResNet introduced the use of residual blocks, which contain skip connections. These connections allow the gradient to flow directly through the network during back propagation, thereby facilitating training of very deep neural networks (Fig. 1). ResNet has been broadly used for computer vision tasks and provides a good baseline benchmark to compare other models and techniques against (He et al., 2016; Shafiq & Gu, 2022).

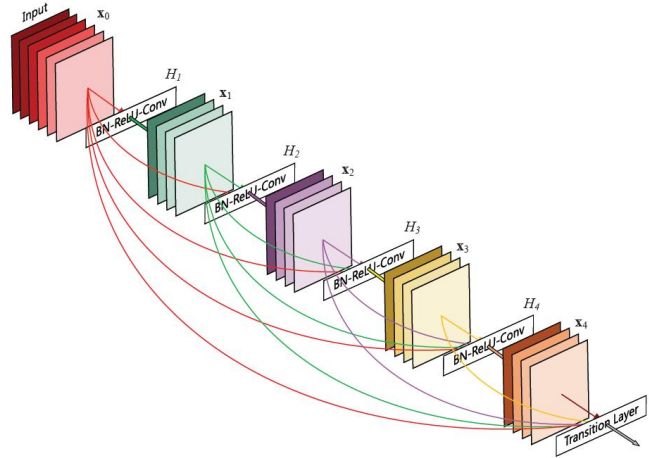

**Figure 2. DenseNet model (https://doi.org/10.1109/Cvpr.2017.243)**

**DenseNet** (Densely Connected Convolutional Networks). DenseNet is a CNN architecture that introduced the concept of dense connections between layers. Whereas prior models only received inputs from its preceding layer, DenseNet constructs direct connections between all layers within a dense block, creating compact models with fewer parameters (Fig. 2). This design facilitates feature reuse, thereby mitigating the vanishing gradient problem (Huang et al., 2017). DenseNet is a popular architecture and offers high accuracy and performance. However, it comes as the cost of being memory-intensive and requiring a high computational cost.

**EfficientNet**, EfficientNet is a breakthrough CNN architecture that was engineered to optimize both accuracy and computational efficiency (Fig. 3). It employs a compound scaling technique that uniformly scales network width, depth, and resolution, achieving higher performance with fewer parameters (Tan & Le, 2019).

**ConvNeXt**, ConvNeXt is a cutting-edge CNN architecture that was engineered with transformer design philosophies in mind. ConvNeXt implements macro design, ResNeXt, inverted bottleneck, large kernel size, and layer-wise micro design. These designs marginally increase the model's performance, culminating in a state-of-the-art model (Liu et al., 2022). ConvNeXt is a modern architecture that can outperform transformers while retaining the straightforwardness of convolutional designs (Liu et al., 2022).

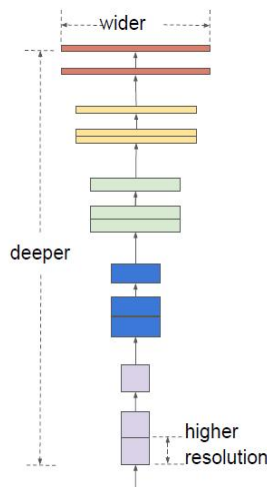

**Figure 3. EfficientNet model (https://doi.org/10.1109/Iccv48922.2021.00986)**

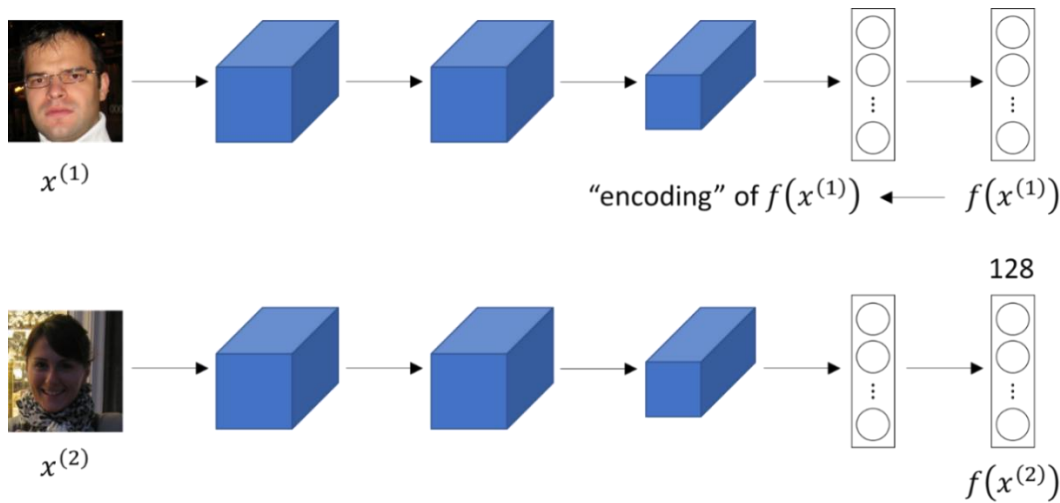

**Figure 4, Siamese network model, (https://doi.org/10.1007/978-1-0716-0826-5_3)**

**Siamese Network**, Siamese networks are a type of neural network architecture that contain two identical convolutional sub-networks. These sub-networks process an input and each return a set of feature vectors. A similarity score between the inputs can be calculated by computing the Euclidean distance between the feature vectors, which can be used for classification (Chicco, 2021)(Fig. 4). Siamese networks are commonly used for tasks like facial recognition and signature verification due to their flexibility and data efficiency. Siamese networks can generalize well, and are able to classify new classes without retraining, while traditional CNNs need to be retrained to handle new classes. Siamese networks require little data after training, only needing one example of each class to classify images in one-shot learning. However, compared to traditional CNNs, Siamese networks can be more difficult to train due to their semi-supervised nature. Siamese networks also require more computation due to their repeated structure and take up more storage due to their need for a support set during classification.

## 3. Methods

**Dataset**

The Cat Individual Images dataset was download from Kaggle (https://www.kaggle.com/datasets/timost1234/cat-individuals/data). The dataset contained images of 518 unique cats, with each class (images of an individual cat) including 6 to 358 images. In total, the dataset included 13536 images. Some images contained multiple cats which would introduce noise to the dataset, so these images were excluded. In addition, we excluded classes with less than 8 images to ensure enough data for training, validation, and testing. After processing, 466 cats with 11076 images were used in the present study. We randomly partitioned the dataset into 8746 train images, 1398 validation images and 932 test images. Each class included 3 images for validation and 2 images for testing, with no duplicates.

**Image Preprocessing**

Most images in the dataset had a super high resolution (2048x2048) and contained a larger space than just the cat. We first used a YOLOv5 model pretrained on the COCO128 dataset to create bounding boxes for cats in each image, including their face and body. Then, we reshaped the bounding boxes to maintain an aspect ratio of 1:1 and added black pixels as padding if the bounding box extended out of the original image. Next, we cropped each image along the bounding boxes to isolate cat information. Last, we used MATLAB to resize the images to 224x224 pixels with 3 channels (RGB) and convert the images to the PNG format (note: the original JPG images couldn't be identified by PIL on the Google Collab servers). 224x224 pixel images with 3 channels were used in most previous studies, and was shown to speed up the training, validation, and testing.

**Augmentation**

The volume of training data is crucial for the training of a deep learning model. The performance on vision tasks increases logarithmically based on volume of training data size (Sun et al., 2017). A small volume of training data can lead to overfitting of the model, damaging its ability to generalize to new data. Data augmentation can provide a method of generating more training data by modifying existing data, creating more training samples and increasing diversity in the dataset, hence increasing performance (Shorten & Khoshgoftaar, 2019). In this paper, we apply a combination of several augmentation methods including flipping, rotation, translation, distortion and information dropping to the training dataset to increase its effective size. The original validation and test sets were retained to evaluate the accuracy of the model. The detailed methods are indicated below:

Flipping, rotation and translation. These are a common approach for many classification tasks, and provide a way to combat a lack in the sample distribution of training images.

Gaussian blur and Gaussian noise. Previous study shows that networks are susceptible to quality distortions in images, particularly to blur and noise (Dodge & Karam, 2016). Applying these augmentations improves the robustness of the model and allows it to generalize for images of less quality.

Color jitter. Color jitter varies brightness, contrast, hue, and saturation in test images. This is a powerful augmentation technique that simulates different lighting conditions and produces diversity within the training dataset.

Cutout. Cutout randomly replaces a rectangular region of an input image with black pixels. This method helps models recognize occluded objects and prevents overfitting.

RandomPerspective. RandomPerspective is a randomly changes the perspective of an input image. This allows the model to be invariant to geometric distortions of input images and recognize objects at different angles.

Normalization. Normalization scales the pixel values of an input image to a certain range, normally 0 to 1. This can help the model converge faster and increases performance by making the input data more consistent and less sensitive to outliers.

**Model Selection**

Many neural networks have been developed in the past decades, such as: ResNet, DenseNet, EfficientNet, ConvNeXt, Siamese networks. These networks have been broadly used in facial recognition. Here, we tested efficiency of these neural networks in individual cat recognition with different models: ResNet-50, DenseNet-121, EfficientNet-B4, ConvNeXt-Tiny, and a Siamese network with a DenseNet-121 backbone.

**Model Training**

We conducted model training on Google Collab with NVIDIA T4 GPUs. Each model was trained for 50 epochs or until the model converged. Model metrics and weights were recorded using Weights & Biases.

**Traditional CNNs**

We trained each model using two methods: fine-tuning and transfer learning. Fine-tuning is a technique where a pretrained model is retrained on a new dataset, allowing the model to perform a task that is related to the original task it was trained on. This method allows for the model to leverage existing knowledge learnt from the original dataset, resulting in faster convergence, and increased accuracy and better generalization. Transfer learning is a technique where the weights of a network are frozen, and only the weights of the final linear classification layer are updated. In the context of CNNs, transfer learning allows the retention of the feature extractors of the original model, training the new model to make predictions based on the extracted features. This method requires less computation and memory during training, results in faster convergence, and increased accuracy and generalization.

We also applied the AdamW Optimizer for all models and set the initial learning rate to 0.01. We selected a ReduceLROnPlateau scheduler with a patience of 5 and a factor of 0.5. This scheduler decreases the learning rate when the model stops improving, allowing for a higher performance ceiling. For the loss function, we applied cross entropy loss, which is commonly used for classification tasks.

**Siamese Network**

We first attempted to train the Siamese network using fine tuning. However, we found that the model improved very slowly. We hypothesized that this was due to the model having to relearn both feature extraction as well as embedding, thus drastically reducing performance. Therefore, we pivoted to solely using transfer learning to train the Siamese network. Here, we applied the AdamW optimizer and set the initial learning rate to 0.005 and 0.0005. We select a StepLR scheduler with a gamma of 10 and a step size of 0.5. For the loss function, we select triplet loss, which is a loss function specially designed for training Siamese networks for tasks such as facial recognition.

The triplet loss function is a loss function that aims to train models learn to embed similar images closely in Euclidean space (Fig. 5). Triplet loss takes a margin value α=1.0, and three feature embeddings of length N=512: an anchor $f(x^a)$, a positive example $f(x^p)$ and a negative example $f(x^n)$.

$$\ell_{\text{triplet}} = ||f(x^a) - f(x^p)||_2 + ||f(x^a) - f(x^n)||_2 + \alpha$$

The loss function minimizes the Euclidean distance between the anchor and the positive while maximizing the distance between the anchor and the negative (Fig. 5).

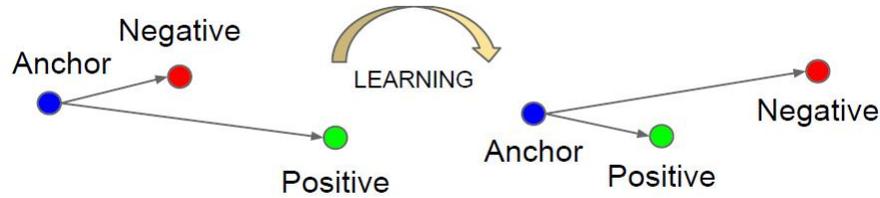

**Figure 5 The triplet loss model (Florian Schroff et al, IEEE Explore, 2015)**

**Siamese Network Classification**

We randomly select a support set of images from the training set, sampling 3 images from each class. These images are embedded by the model into embeddings of length N, of which we take the mean. Each validation or test image is also embedded to an embedding of length N.

The K-Nearest Neighbor (KNN) algorithm uses the Euclidean distance between feature vectors in a N-dimensional space to classify embeddings. We fit the KNN with the support set embeddings as well as labels. The query vector is introduced and classified based on the classes of the K nearest points to the query vector.

## 4. Results

### 4.1 Training and Validation

To compare the efficiency of different neural networks in individual cat recognition, we first trained the ResNet, EfficientNet, DenseNet and ConvNeXt models with the fine-tuning method. We found the accuracy of the ResNet improved at a moderate rate in fine-tuning. After convergence, it showed signs of heavy overfitting, with a maximum difference of 17.9% between train and validation accuracy (Fig. 6). ResNet accomplishes a maximum accuracy of 71.2% in validation (Fig. 6D). EfficientNet showed drastic overfitting in the first few epochs, with a maximum difference of 14.7% between train and validation accuracy. However, the model displayed steep improvement of 65.9 % in validation accuracy in the $7^{th}$ epoch, likely due to a decrease in learning rate. EfficientNet showed a significant improvement compared to ResNet in the validation set, achieving a maximum accuracy of 79.9%. However, EfficientNet also showed heavy overfitting after convergence, with a maximum difference of 14.7% between train and validation accuracy (Fig. 6). DenseNet's training curves showed similar characteristics to ResNet, and achieved similar performance in the validation set as EfficientNet, achieving a maximum accuracy of 81.0% (Fig. 6). ConvNeXt yielded bad results during fine-tuning, showing a static validation accuracy of 0.2% (Fig. 6). Train accuracy fluctuated early on, but also stagnated at 1.551%.

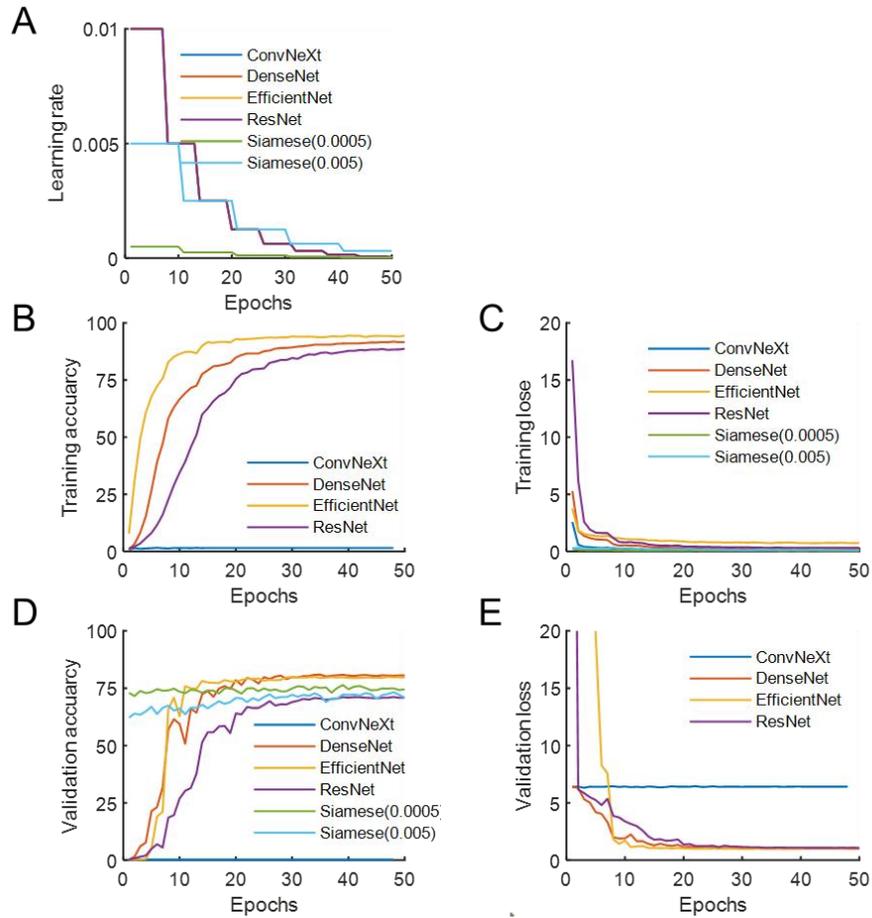

**Figure 6: the comparation of training and validation accuracy using the fine-tuning method and Siamese networks.** A. the learning rate of the 5 models during training. B and C, the training accuracy and loss of ConvNeXt, DenseNet, EfficientNet, ResNet, and Siamese network models. D and E, the validation accuracy and loss of ConvNeXt, DenseNet, EfficientNet, ResNet, and Siamese network models.

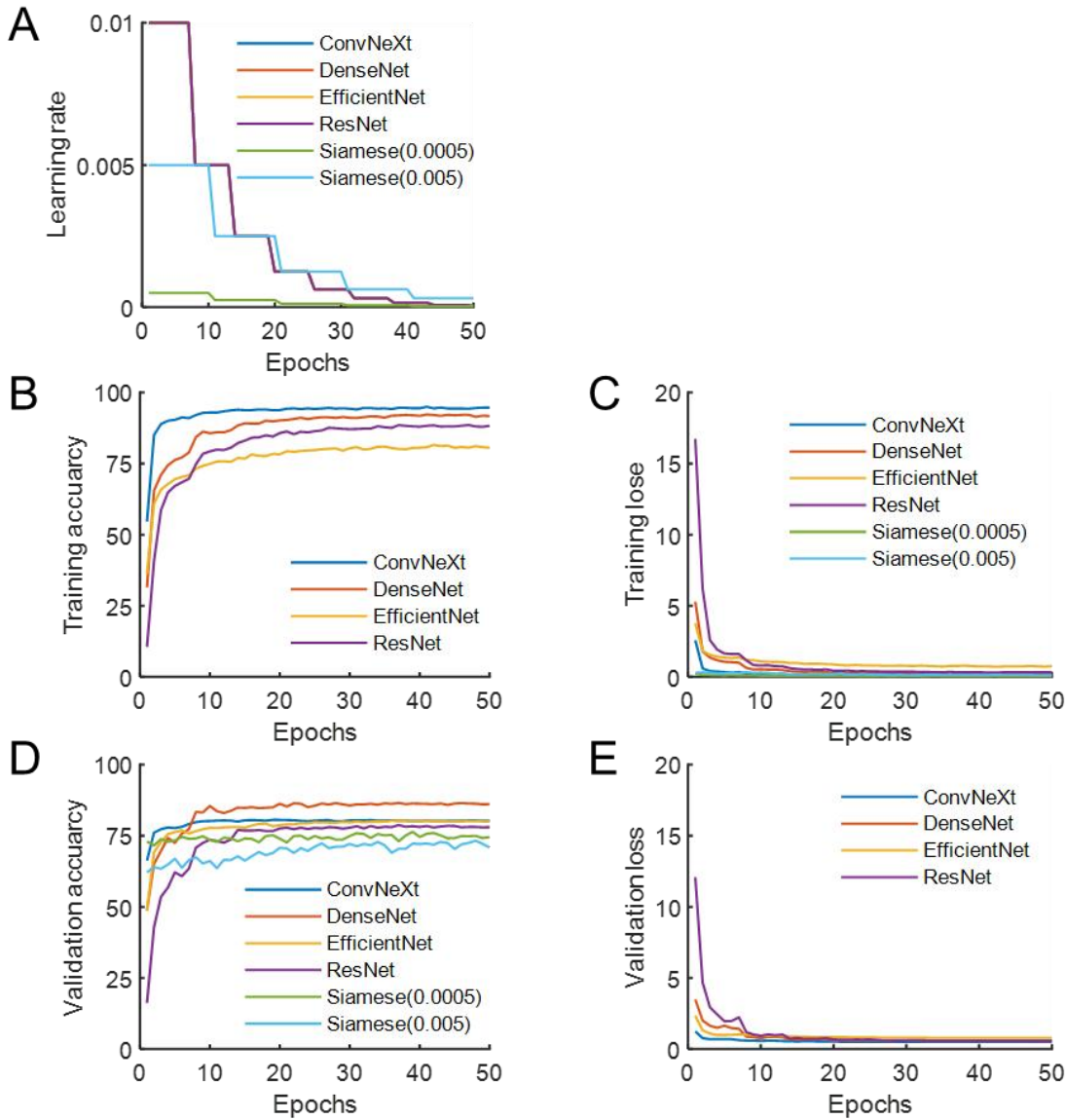

**Figure 7: the comparation of training and validation accuracy using the transfer learning method and Siamese networks.** A. the learning rate of the 5 models during training. B and C, the training accuracy and loss of ConvNeXt, DenseNet, EfficientNet, ResNet, and Siamese network models. D and E, the validation accuracy and loss of ConvNeXt, DenseNet, EfficientNet, ResNet, and Siamese network models.

Then, we compared our results with ResNet, EfficientNet, DenseNet, and ConvNeXt trained with transfer learning, as well as with the Siamese networks. We found that all models had a higher starting point for validation accuracy during the first epoch (Fig. 7). All traditional CNN models converged within 15 epochs, and achieved higher validation accuracies compared to fine-tuning (Fig. 6D and 7D). ResNet showed significantly higher validation accuracy during transfer learning, with a difference of about 7.6%. The validation accuracy achieved a maximum of 78.8%, similar to that of EfficientNet with the fine-tuning approach. EfficientNet showed very marginal improvement compared with fine tuning, only achieving a difference of around 0.5% in validation accuracy. Interestingly, Train accuracy decreased significantly for EfficientNet during transfer learning, with a decrease of around 12.9%, resulting in almost no overfitting (Fig. 7).

ConvNeXt had a very high starting point at 66.3% validation accuracy during transfer learning, and quickly converged to around 80% accuracy, showing much more positive results compared to fine-tuning (Fig. 7). DenseNet trained with transfer learning achieved the highest performance of all models, with a maximum validation accuracy of 86.6%. Similar to ResNet, DenseNet's train accuracy plateaued at similar accuracy compared to the fine-tuning approach (Fig. 7).

Siamese networks are usually thought to be a better approach for facial recognition. However, the Siamese networks we trained showed very gradual improvement in validation accuracy with large fluctuations. This improvement generally coincided with a decrease of the learning rate. However, the model showed no significant improvements with a learning rate below 0.000125. In addition, the maximal validation accuracy only reached 76.4% and 73.3% for Siamese (0.0005) and Siamese (0.005), which is less than EfficientNet and DenseNet trained with the fine-tuning method (Fig. 6D and 7D).

### 4.2 Testing

After training, each model was tested on the test dataset. The accuracy of each model is shown in Table 1.

**Table 1: Validation and test accuracy of different neural networks in individual cat recognition.**

| Neural networks | Validation | | | Testing | | |
| --- | --- | --- | --- | --- | --- | --- |
| | | fine-tune | transfer | | fine-tune | transfer |
| **ResNet50** | X | 71.2 | 78.8 | X | 72.6 | 78.5 |
| **DenseNet** | X | 81.0 | 86.6 | X | 74.2 | 79.6 |
| **EfficientNetB4** | X | 79.9 | 80.4 | X | 77.0 | 77.4 |
| **ConvNeXt** | X | 0.2 | 80.6 | X | 20.8 | 85.2 |
| **Siamese (0.005)** | 73.3 | X | X | 66.2 | X | X |
| **Siamese (0.0005)** | 76.4 | X | X | 69.7 | X | X |

X, data does not exist.

We found that all traditional CNNs have better performance in transfer learning than fine-tuning in both validation and testing (Table 1). In addition, ConvNeXt using transfer learning performed surprisingly well on the test set, with a 4.7% increase in accuracy compared to the validation set (Table 1). However, the Siamese networks only reached 66.2% accuracy during testing (Table 1). So, transfer learning yields higher accuracy during validation and testing for individual cat recognition. We expected the Siamese networks to perform the best. However, it showed worse performance and may require further testing in future.

## 5. Discussion

In the present study, we first used YOLOv5 to extract cats' bodies as well as their faces from source images to use in later model training. This is opposed to prior research, which only use facial images in tandem of human facial recognition. This manipulation allows the networks to observe unique fur patterns, which may be important markers for identification. In addition, we compared performance between traditional CNNs and Siamese networks in accomplishing this task. Our results showed that traditional CNNs trained with transfer learning have better performance than models trained with fine-tuning or Siamese networks in individual cat recognition.

ConvNeXt showed peculiar results. During fine-tuning, ConvNeXt showed anomalous static performance, with a constant validation accuracy of 0.1953%. We hypothesized that this was due to all of the model's gradients becoming zero. However, during testing, ConvNeXt surprisingly achieved an accuracy of 20.833%, which doesn't support this hypothesis. In addition, ConvNeXt trained with transfer learning showed a 4.7% increase in accuracy in testing compared to validation, whereas most other models displayed a decrease in performance. This may imply that the ConvNeXt feature extractor performs better at generalization than the other models. Or, the model may have underperformed in the validation set. ConvNeXt approaches CNNs with design philosophies from transformers, which could be the cause of this unexpected behavior. We plan on further investigation to find the exact reason for these results.

Siamese networks are usually thought to be a better approach for human facial recognition (Fouladi-Ghaleh, 2020; Schroff et al., 2015). Their ability to classify novel classes without retraining makes them suitable for deployment in low powered devices which lacking processing power. However, although it showed a trend of increased validation accuracy with decreasing learning rate, the Siamese network performed significantly worse on the test set compared to the validation set and also worse than traditional neural networks. These results may be due to a low variance in the train set which affecting the model's ability to generalize during the testing. In addition, for the Siamese networks, we classified query images with a KNN algorithm where K=1, taking the mean of embeddings of the support set. Increasing the K value and fitting the support set without taking the average could result in more accurate classification. However, this would not affect the ability of the Siamese network to learn to embed images, and would only result in a marginal increase in validation and test accuracy. We plan to conduct further testing with different parameters and datasets to evaluate Siamese networks' efficacy in this task.

There are some limitations to our study. 1) The small dataset. We selected the Cat Individual Images dataset from Kaggle, which contains 13536 images of 518 unique cats. This dataset was considerably small compared to image sets used in other research papers. For example, the VGGFace2 dataset is made of around 3.31 million images divided into 9131 classes. A small dataset could have significantly increased the overfitting of our models during training and hamper our models' ability to generalize. 2) Our dataset contained multiple instances of similar images. The cats in many image sets are in the same location or in similar poses, which introduces a source of bias. The lack of variability could have resulted in a simple learning task, resulting in worse generalization. 3) The pretrained models we selected were all pretrained on ImageNet, which might pose a large source of bias. These pretrained models might be biased to recognize features that are more common in ImageNet, and partially rely on objects in the background to classify images. In the future, we plan to expand our dataset using images of cats gathered from online social platforms such as Instagram and images of cats taken in pet stores or in the wild. This will introduce more variability into the dataset and improve the general application of our research. We will also explore methods to reduce bias from pretrained models, such as training models from scratch or using images with a variety of backgrounds for the same cat.

## 6. Conclusion

In present study, we found that traditional CNNs trained with transfer learning have better performance than models trained with a fine-tuning method or Siamese networks in individual cat recognition. In addition, ConvNeXt and DenseNet yield significant results which could be further optimized for cat recognition in the wild and in pet stores.